\documentclass[10pt,twocolumn,letterpaper]{article} 

\usepackage{avss}
\usepackage{times}
\usepackage{epsfig}
\usepackage{graphicx}
\usepackage{amsmath}
\usepackage{amssymb}
\usepackage{multirow}
\usepackage{amsfonts,amssymb}
\usepackage{amssymb}
\usepackage{makecell,multirow,diagbox,subfigure,amsfonts,bbding,verbatim,amsmath, bm}
\usepackage{graphicx}
\usepackage{float}

\usepackage[pagebackref=true,breaklinks=true,letterpaper=true,colorlinks,bookmarks=false]{hyperref}

\avssfinalcopy 

\ifavssfinal\fi

\begin{document}

\title{Refining Action Boundaries for One-stage Detection}

\author{Hanyuan Wang\and Majid Mirmehdi\and Dima Damen\and Toby Perrett\\
Department of Computer Science,
University of Bristol, Bristol, U.K.\\
{\tt\small \{hanyuan.wang, dima.damen, toby.perrett\}@bristol.ac.uk, majid@cs.bris.ac.uk}
}

\maketitle
\thispagestyle{empty}

\begin{abstract}

Current one-stage action detection methods, which simultaneously predict action boundaries and the corresponding class, do not estimate or use a measure of confidence in their boundary predictions, which can lead to inaccurate boundaries.
We incorporate the estimation of boundary confidence into one-stage anchor-free detection, through an additional prediction head that predicts the refined boundaries with higher confidence.
We obtain state-of-the-art performance on the challenging EPIC-KITCHENS-100 action detection as well as the standard THUMOS14 action detection benchmarks, and achieve improvement on the ActivityNet-1.3 benchmark.

\end{abstract}

\let\thefootnote\relax\footnote{978-1-6654-6382-9/22/\$31.00 ©2022 IEEE}

\section{Introduction}
\label{sec:intro}


Current video understanding approaches \cite{slowfast, I3D, TSN} recognise actions on short, trimmed videos. These assume the boundaries of actions are already given, and thus focus on the class prediction problem solely. However, most real-life videos are untrimmed and contain irrelevant visual content. 
Temporal action detection aims to temporally locate the boundaries of actions and classify them in longer, unscripted and untrimmed videos \cite{EPIC100, Ego4d, THUMOS14, Activitynet}, which is crucial for video analysis.

Two-stage action detection approaches, such as \cite{SSN, BSN, BCGNN, BMN, DCAN, BSN++}, were built on top of successful recognition models  \cite{slowfast, duan2022revisiting, xu2022cross} and widely used as reference methods on simple action detection baselines  \cite{THUMOS14, Activitynet}. They first generate candidate proposals based on pre-defined sliding windows or matching locations with high probabilities scores, and then classify them to obtain the final predictions. 
However, such two-stage methods are inefficient for the wider variety of actions, action lengths and action/background densities found in longer untrimmed videos, since a large number of redundant candidate proposals are produced by sliding windows and location matching. 

\begin{figure}[t!]
	\centering
	\includegraphics[width=0.9\linewidth]{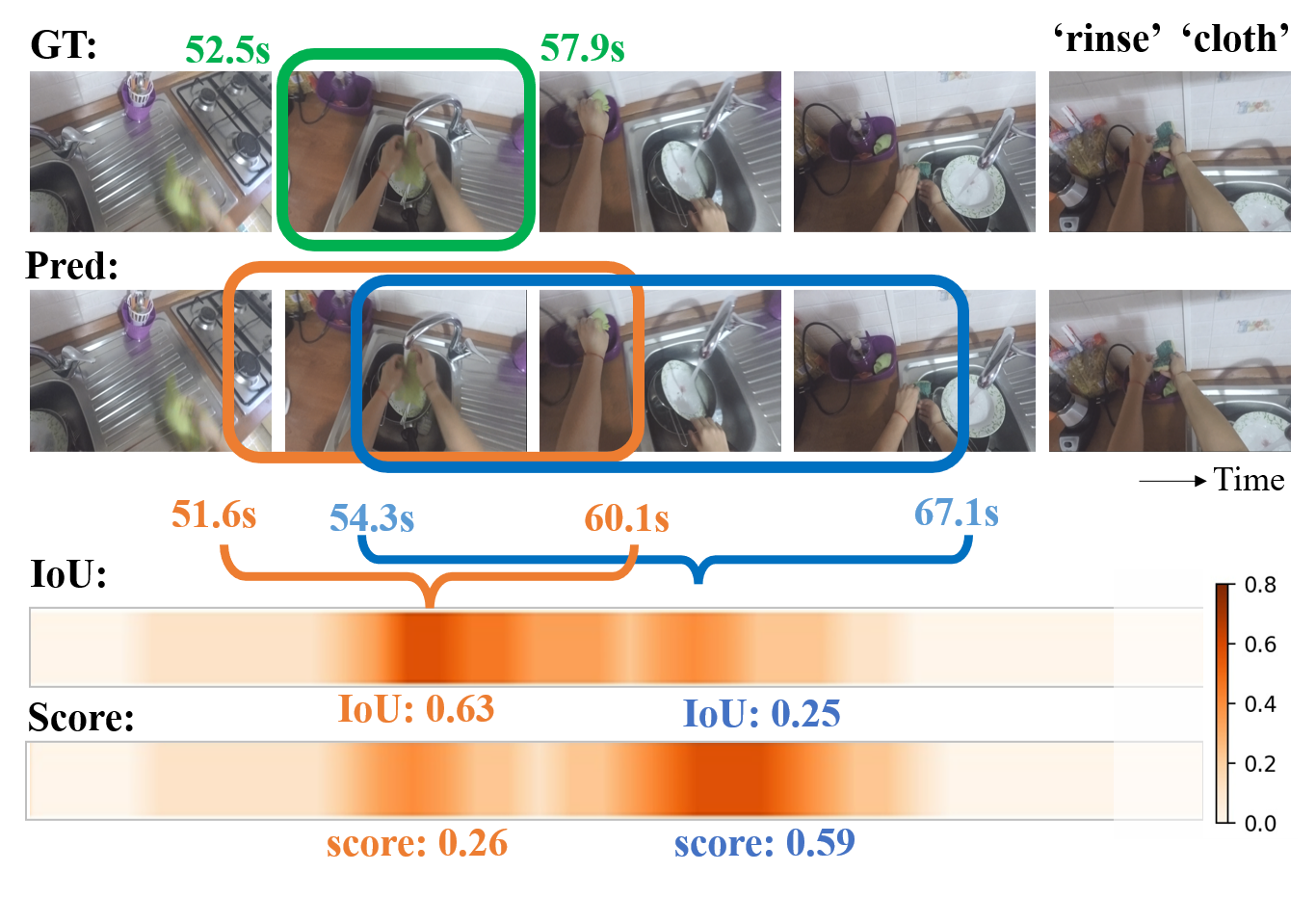}
	\vspace*{-5pt}
    \caption{An illustration of the misalignment between the value of tIoU and classification scores of predicted proposals, caused by the absence of boundary confidences. 
    \textcolor[RGB]{50,205,50}{Green} denotes the ground truth, \textcolor[RGB]{0,113,199}{blue} and \textcolor[RGB]{255,97,0}{orange} denote predictions produced by ActionFormer~\cite{actionformer}.
    Specifically, when the boundary confidence is not considered as the ranking metric, the prediction with a higher classification score but poor boundaries (\textcolor[RGB]{0,113,199}{blue}) is chosen, rather than the prediction with better boundaries (\textcolor[RGB]{255,97,0}{orange}).
}
	\label{fig: motivation}
\end{figure}

More recently, one-stage methods have been proposed, where the network simultaneously predicts the current action for each timestep and its associated boundaries \cite{A2Net, AFSD, actionformer}. 
In this paper, we show that these methods are missing the boundary confidence in proposal regression and evaluation.
This can lead to imprecise localisation due to insufficient boundary information, especially in the case of actions of various lengths found in egocentric data, such as EPIC-KITCHENS\cite{EPIC100}. An example of the action `rinse cloth' from~\cite{EPIC100} is shown in Figure \ref{fig: motivation}, where the prediction with a higher classification score has a lower overlap between boundaries and ground truth (blue), while the prediction with better boundaries has a lower classification score (orange). This is due to the absence of boundary confidences, resulting in poor regression and unreliable scores.

In this paper, we consider the extent of an action proposal and estimate the confidence of the \textit{start} and \textit{end} frames of the action segment, jointly.  We supervise the confidence from the relative distance between the estimated frame and the ground truth frame, for both the start and end boundaries of the action. This confidence information is leveraged to refine the boundaries of proposals which leads to state-of-the-art action detection results on EPIC-KITCHENS-100 \cite{EPIC100} and THUMOS14 \cite{THUMOS14}, and improvement on ActivityNet-1.3 \cite{Activitynet}.

In summary, we introduce a boundary head for one-stage anchor-free action detection which estimates boundary confidence scores based on relative distances. We obtain state-of-the-art results on EPIC-KITCHENS-100 and THUMOS14 action detection, using the same backbone as the current state-of-the-art. Notably, significant improvement is achieved on EPIC-KITCHENS-100, which indicates that our method performs well on complex actions of various lengths. Further, we provide detailed ablations, including investigating confidence scores and the effect of action lengths.


\section{Related Work}
\label{sec: related work}
Action detection methods can be grouped into two categories: two-stage and one-stage. 

\noindent {\bf Two-stage action detection:}
Two-stage methods 
first generate a set of candidate proposals and then classify each proposal.  
They typically generate proposals by pre-defined sliding window and grouping temporal locations with high probabilities of being within an action \cite{SSN} or close to a boundary \cite{BSN, BCGNN}. Action and boundary combinations can be selected based on high boundary confidence \cite{BMN, DCAN}, or a combination of separately calculated boundary and action scores \cite{BSN++}. This generation process can struggle when presented with sequences containing many dense actions of varying lengths, such as EPIC-KITCHENS-100 \cite{EPIC100}.

\noindent {\bf One-stage action detection:} One-stage methods 
improve detection efficiency by simultaneously predicting action proposals and their associated classes. 
One approach is to generate candidate boundaries by modelling temporal relationships \cite{Turntap, SSAD, GTAN, RC3D, PBRNet}.
However, these methods rely on pre-defined anchors, causing them to struggle when presented with a wide range of action durations.
Inspired by the DETR framework \cite{DETR} for object detection, some works use learned action \cite{TadTR} or graph \cite{AGT} queries as input to a transformer decoder. Whilst a promising direction, these methods are not suitable for long videos due to attention scaling issues.
Anchor-free methods \cite{A2Net, AFSD, actionformer} simultaneously predict classification scores and a pair of relative distances to boundaries for each timestep. 

Recently, ActionFormer \cite{actionformer} generated these predictions with a multi-scale transformer encoder to model both short- and long-range temporal dependencies, with simple classification and boundary regression heads, and achieved state-of-the-art results on a number of benchmarks. 
In this work, we adopt the same multi-scale transformer encoder and pipeline as ActionFormer \cite{actionformer}, but incorporate the ability to estimate boundary confidences. 

\section{Method}
\label{our method}
We first briefly review ActionFormer~\cite{actionformer}, and then introduce our novel boundary head, which is incorporated into ActionFormer to achieve better performance.

\subsection{Overview of ActionFormer}
ActionFormer first extracts a feature pyramid based on local self-attention, and then uses light-weight heads to simultaneously predict classification scores and a pair of relative distances to boundaries for each timestep.

\noindent\textbf{Transformer-based feature pyramid:}
ActionFormer extracts features from an untrimmed sequence and passes them to a multi-scale transformer encoder \cite{actionformer} to construct a feature pyramid sequence. The feature pyramid sequence contains multiple resolutions for each timestep, which allows a single timestep to detect short and long actions. 

\noindent\textbf{Prediction heads:}
A classification head uses the feature pyramid sequence to predict action labels and classification scores for each timestep in multiple resolutions, and similarly, a regression head predicts relative distances to the predicted start and end boundary locations, for every timestep in the feature pyramid.

\noindent\textbf{Training and Inference:}
The network is trained by minimizing the multi-part losses of the classification head and the regression head. 
For the classification head, a focal loss~\cite{focal} is used to balance loss weights between easy and hard examples.
For the regression head, it minimises the distance between the ground truth boundaries and the predicted boundaries using the GIoU loss \cite{iouloss}. 
At inference, they predict a pair of relative distances to boundaries and a classification score to give a proposal for each timestep across all pyramid levels. These candidate proposals are ranked by classification scores and further filtered to obtain the final outputs of actions.

\subsection{Boundary Head}
\label{boundary head}

In ActionFormer, the regression head nominates where the boundaries are, without providing any confidence of the locations as boundaries. To address this, we compute the boundary confidence at the same time as the boundary location prediction. One approach could be using a separate branch to directly predict boundary confidence. However, this may lead to learning conflicts in the anchor-free pipeline, where the original network must learn the relative distances between the current temporal location and ground truth boundaries, rather than the confidence that the current location is a boundary (demonstrated in Section \ref{abaltion}). 

\begin{figure}[t]
	\centering
	\includegraphics[width=0.99\linewidth]{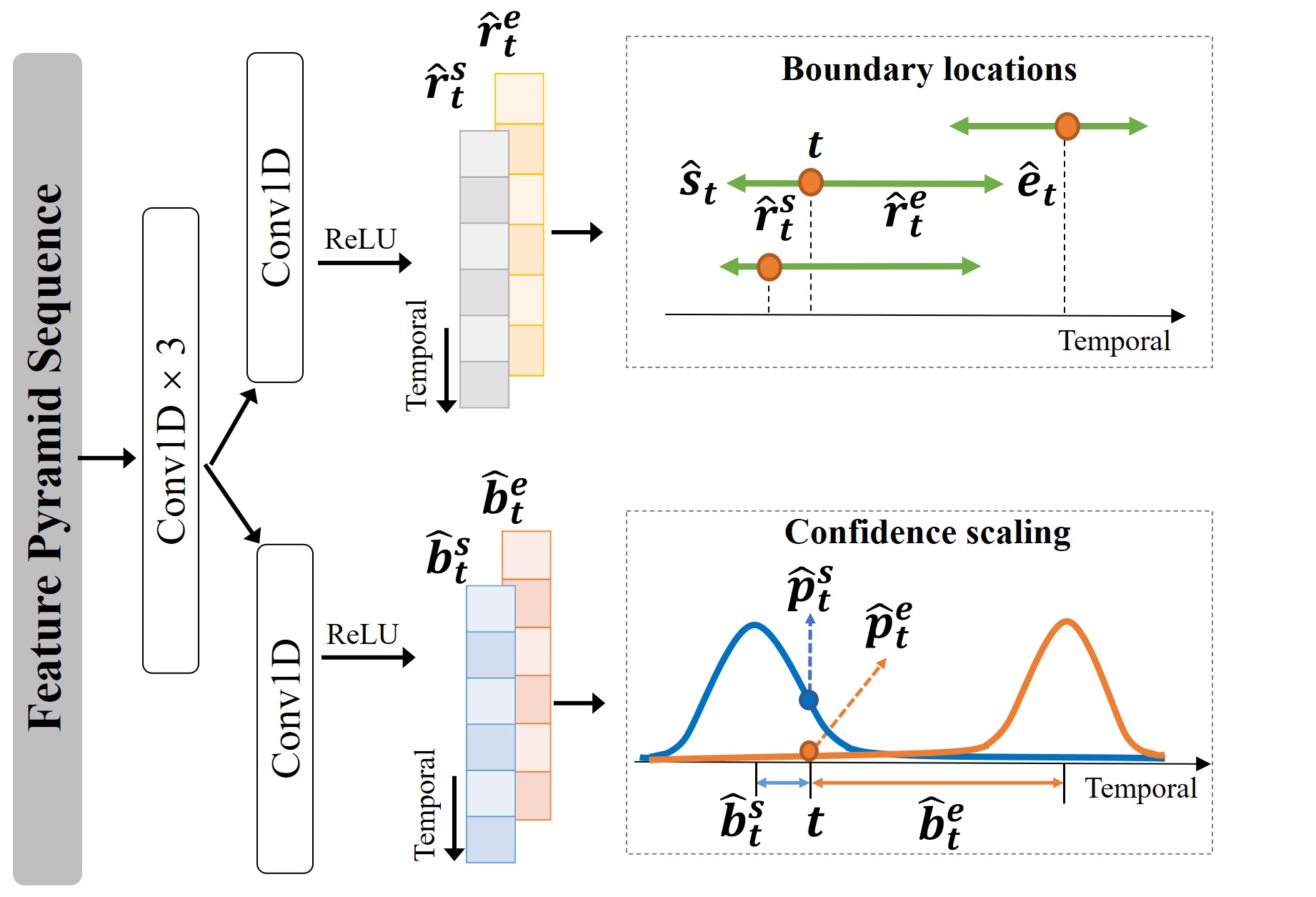}
	\vspace*{-5pt}
    \caption{Overview of our boundary head. Taking in the feature pyramid sequence, two branches share most of the weights but have separate top layers. The first branch predicts start and end boundary \textbf{locations}. The second branch predicts boundary \textbf{confidence}.
    A confidence scaling encodes the assumption that a location $t$ can be more confident about boundaries which are closer.}
	\label{fig: main framework}
\end{figure}

We design a simple but effective boundary head, which computes boundary probabilities via a confidence scaling, weighted by how close boundaries are. This approach, instead of directly producing boundary probabilities, means the network is better able to keep the consistency of two parallel branches in predicting and optimizing and make the learning process easier.
As shown in Figure \ref{fig: main framework}, the boundary head takes in the feature pyramid sequence, and contains two branches. 
These two branches share most of the weights, 
as there is shared information between the two tasks (e.g. high start confidence and small relative start offset are related), 
but they have separate top layers.
The first branch predicts relative distances $(\hat{r}_{t}^{s},\ \hat{r}_{t}^{e})$ to the start and end boundaries for each location $t$ in the feature pyramid~\cite{AFSD, actionformer}. Thus, the corresponding predicted start and end \textit{locations} are obtained as $\hat{s_t} = t - \hat{r}_{t}^{s}$ and $\hat{e_t} = t + \hat{r}_{t}^{e}$.
The second branch predicts start and end confidence tokens $(\hat{b}_{t}^{s},\ \hat{b}_{t}^{e})$ that are further fed into scaling processing (will be introduced next) to calculate the boundary \textit{confidence} that location $t$ is either an action start or an action end. 

\noindent\textbf{Confidence scaling:}
To assign a confidence that location $t$ is a start (${\hat p}_{t}^{s}$) or end (${\hat p}_{t}^{e}$) boundary, the confidence tokens are weighted such that 
\begin{equation}
  \hat{p}_{t}^{s} = \text{exp} \left({-(\hat{b}_{t}^{s})^{2}/2\sigma^{2}} \right) \  \mbox{and} \  
  \hat{p}_{t}^{e} = \text{exp} \left({-(\hat{b}_{t}^{e})^{2}/2\sigma^{2}} \right) ~.
  \label{eq:gaussian}
\end{equation}
Here, $\sigma$ is a scaling hyperparameter, determined experimentally in Section \ref{abaltion}. Scaling in this manner encodes the assumption that a location $t$ can be more confident about boundaries which are closer. This has previously been explored to improve online detection \cite{Gaussian, ddlstm}, but not for offline use cases.

\noindent\textbf{Label assignment:}
We require ground truth confidence values as supervision signals. We omit all locations where the GIoU between the ground truth and the predicted boundary locations is less than a threshold $\beta$.  We then adopt the BSN approach \cite{BSN} and denote the ground truth start confidence for location $t$ as ${p}_{t}^{s}$.
{Specifically, we define a region of length 1 timestep centered on location $t$, and calculate ${p}_{t}^{s}$ as its overlap ratio to a region of length $d/5$ centered on the ground truth start location, where $d$ is the length of the ground truth action.}
This can be thought of as the value at location $t$ of a start probability curve \cite{BSN, BMN, bottom-up}, and is visualised in Figure \ref{fig: main framework}. In the case of multiple overlapping ground truth start locations, we take the maximum single overlap ratio. The ground truth end confidence ${p}_{t}^{e}$ is calculated similarly.

\noindent\textbf{Training:}
We minimise the difference between the ground truth and predicted confidences using the following two loss functions: 
\begin{equation}
\begin{aligned}
{L}_{conf}^{s}=\frac{1}{T}\sum_{t} (\hat{p}^{s}_{t}-p^{s}_{t})^{2} \ \mbox{and} \  
{L}_{conf}^{e}=\frac{1}{T}\sum_{t} (\hat{p}^{e}_{t}-p^{e}_{t})^{2},
\end{aligned}
\end{equation}
where $T$ is the total number of locations used for training from all levels of the feature pyramid sequence.

In an end-to-end manner, we incorporate our confidence losses into the total loss and optimize a weighted combination of all the losses:
\begin{equation}
\label{total loss}
L_{total} = L_{cls} + \gamma L_{GIoU} + \omega (L_{conf}^s + L_{conf}^e) ~,
\end{equation}
where $L_{cls}$ and  $L_{GIoU}$ are losses for classification and regression and are the same as in ActionFormer.

\subsection{Post-processing}
\label{Inference}
The network produces a proposal from each location $t$ in the feature pyramid containing a predicted start boundary $\hat{s_t}$ and a predicted end boundary $\hat{e_t}$. It also gives an action $\hat{a_t}$ and an action confidence score $\hat{p}_{t}^{a}$ taken from the logits. 
We can obtain the boundary confidences for this proposal by looking up the start boundary confidence at location ${\hat{s_t}}$ and the end boundary confidence at location ${\hat{e_t}}$. These confidences are $\hat{p}_{\hat{s_t}}^{s}$ and $\hat{p}_{\hat{e_t}}^{e}$.

We first multiply the start and end confidences to get the boundary confidence, then combine it with the action confidence to give a single final confidence for the proposal:
\begin{equation}
\label{conf score equation}
\hat c = \hat{p}^{a}_{t} \sqrt{\hat{p}^{s}_{\hat{s}}  \hat{p}^{e}_{\hat{e}}} ~,
\end{equation}
where a square root is used to balance the contributions of boundary and action confidences (demonstrated in Section \ref{abaltion}).
Finally, we follow standard practice \cite{BSN, BMN, BSN++, actionformer} and suppress redundant proposals with Soft-NMS \cite{softnms} to obtain a final set of $M$ predictions $\hat{\Phi}=\left\{(\hat{s},\ \hat{e},\ \hat{c},\ \hat{a})_m \right\}^M_{m=1}$.

\begin{table*}[t]
\scriptsize
\begin{center}
\begin{tabular}{|l|c|c|cccccc|}
\hline
\multicolumn{1}{|c}{\multirow {2}{*}{\bf Method}} &\multicolumn{1}{|c}{\multirow {2}{*}{\bf Venue}} &\multicolumn{1}{|c}{\multirow {2}{*}{\bf Feature}} &\multicolumn{6}{|c|}{\bf mAP@IoU}\\
\cline{4-9}
\multicolumn{1}{|c}{}&\multicolumn{1}{|c}{}&\multicolumn{1}{|c}{}&\multicolumn{1}{|c}{0.1} & 0.2 & 0.3 & 0.4 & 0.5 & Avg. \\ 
\hline\hline
\multirow{1}{*}{BMN \cite{BMN,EPIC100}} &IJCV 2022 &SF \cite{slowfast} &6.95 &6.10 &5.22 &4.36 &3.43 &5.21\\ 
\multirow{1}{*}{AGT \cite{AGT}}  &{arXiv 2021} &I3D \cite{I3D} &7.78 &6.92 &5.53 &4.22 &3.86 &5.66\\
\multirow{1}{*}{TSN \cite{BMN, TADA}} &ICLR 2022 &TSN \cite{TSN} &10.24 &9.61 &8.94 &7.96 &6.79 &8.71\\
\multirow{1}{*}{OWL \cite{OWL}} &arXiv 2022 &SF(A,V)\cite{slowfast, SFA} &11.01 &10.37 &9.47 &8.24 &7.26 &9.29\\
\multirow{1}{*}{TAda2D \cite{BMN, TADA}} &ICLR 2022 &TAda2D \cite{TADA} &15.15 &14.32 &13.59 &12.18 &10.65 &13.18\\
\multirow{1}{*}{AF~\cite{actionformer}$^*$} &ECCV 2022 &SF \cite{slowfast} &\underline{18.02}  &\underline{17.41}  &\underline{16.44}  &\underline{15.17}  &\underline{13.23}  &\underline{16.05} \\
\multirow{1}{*}{Ours} &- &SF \cite{slowfast} &\textbf{19.19} &\textbf{18.61} &\textbf{17.47} &\textbf{16.30} &\textbf{14.33} &\textbf{17.18} \\
\hline
\end{tabular}
\end{center}
\caption{Comparative results on the the EPIC-KITCHENS-100 validation set for the action task (i.e. predict both verb and noun). $^*$AF or ActionFormer only provides results on verb and noun detection separately, so we produce results by modifying it with our multitask classification head. \textbf{Bold} for best model and \underline{underline} for second best.
}
\label{tab:results_epic}
\end{table*}

\section{Experiments}
 
\subsection{Setup}
\noindent {\bf Datasets:}
We evaluate our method on three datasets: EPIC-KITCHENS-100~\cite{EPIC100}, ActivityNet-1.3~\cite{Activitynet} and THUMOS14~\cite{THUMOS14}.
Two of the most widely used baselines for action detection are ActivityNet-1.3 \cite{Activitynet} and THUMOS14 \cite{THUMOS14}. ActivityNet-1.3 contains 19,994 videos with 23,064 instances of 200 action classes, at an average of 1.5 instances per video. THUMOS14 consists of 413 videos with 38,690 instances of 20 action classes, and an average of 15.4 instances per video. 

More recently, interest in wearable cameras has led to the collection and annotation of large-scale egocentric datasets, such as EPIC-KITCHENS-100 \cite{EPIC100}, which consists of 700 videos with 89,977 verb/noun action instances. There are 96 verb and 300 noun classes, with 4,053 complicated action instances (e.g. verb/noun pairs such as ``pickup fork'' rather than ``cycling''), with much longer sequences containing an average of 128 action instances per video. They also contain a wider range of action durations, along with significant overlap between actions due to participants acting in a natural manner in familiar environments. These provide a much more difficult test for action detection methods \cite{EPIC-challenge-2021}. 

\noindent {\bf Evaluation Metrics:}
Following the official settings, we use mean Average Precision (mAP) at different Intersection over Union (tIoU) thresholds to evaluate the performance of action detection. 
The mAP is the average precision across all action classes.
On EPIC-KITCHENS-100, the tIoU thresholds are set to \{0.1, .., 0.5\} at step size of 0.1. On ActivityNet-1.3 they are \{0.5, 0.75, 0.95\}, and on THUMOS14 they are \{0.3, .., 0.7\} at step size of 0.1.

\noindent {\bf Baselines:}
We compare against state-of-the-art (SOTA) approaches on EPIC-KITCHENS-100~\cite{BMN, AGT, TSN, OWL, TADA, actionformer}, ActivityNet-1.3 and THUMOS14~\cite{BSN, BMN, GTAD, BCGNN, BSN++, TCANET, TVNet, DCAN, RCL, RefactorNe, MUSES, SSN, GTAN, TadTR, AFSD, TALLFormer, MRBD} benchmarks.
As ActionFormer only provides results and code for separate verb and noun detections (not combined action detection which is standard practice) on EPIC-KITCHENS-100, we modify it with our classification head to provide comparable results. We also provide separate verb and noun detection results of our own to compare.

\noindent {\bf Implementation details:}
For feature extraction and pyramid generation, we follow ActionFormer \cite{actionformer} for all datasets. Briefly, we use SlowFast \cite{slowfast} features for EPIC-KITCHENS-100, TSP \cite{TSP} features for ActivityNet-1.3 and I3D \cite{I3D} for THUMOS14. The transformer encoder generates a feature pyramid with 6 levels, with a level scaling factor of 2. The length of the first pyramid level is 2304 for EPIC-KITCHENS-100 and THUMOS14, and 768 for ActivityNet-1.3.

For label assignment and training, we find the ground truth action that location $t$ is in and take its start $s^*$ and end $e^*$ boundaries. When multiple ground truth regions overlap with $t$, only the shortest ground truth region is used to make regression during training easier. Following \cite{fcos, bridging, actionformer}, if a location $t$ is not in an action, or is too close to a ground truth action boundary (i.e. not in the middle $\alpha$ timesteps of an action), it is omitted from the loss calculation. The value of $\alpha$ is set as 3.
In Equation \ref{eq:gaussian}, we select $T$ samples to train the network. Specifically, we set the GIoU threshold for rejecting proposals used to train boundary confidence prediction as $\beta=0.5$.  
For the total loss in Equation \ref{total loss}, the weights are set as $\gamma=0.5$ and $\omega=0.5$. These are the same for all datasets. On EPIC-KITCHENS-100, the classification loss weight for verb/noun is set as $0.5$.

In the classification head, ActionFormer just returns logits and predicts actions for a single action task (as in ActivityNet-1.3 and THUMOS14). For the compound actions found in EPIC-KITCHENS-100, which are verb/noun pairs, it is not practical to return confidences for every possible verb and noun combination. Instead, we take the top-$v$ verb and top-$n$ noun predictions. The candidate actions are every combination of top-$v$ verbs and top-$n$ nouns, and their confidences are the verb and noun logits multiplied together. Our main results use $v=10$ and $n=30$.

\begin{table*}[t]
\scriptsize
\begin{center}
\begin{tabular}{|l|l|cccccc|}
\hline
\multicolumn{1}{|c}{\multirow {2}{*}{\bf Task}}&\multicolumn{1}{|c}{\multirow {2}{*}{\bf Method}} &\multicolumn{6}{|c|}{\bf mAP@IoU}\\
\cline{3-8}
\multicolumn{1}{|c}{}&\multicolumn{1}{|c}{}&\multicolumn{1}{|c}{0.1} & 0.2 & 0.3 & 0.4 & 0.5 & Avg.\\ 
\hline\hline
\multirow{2}{*}{Verb} &G-TAD \cite{GTAD,actionformer} &12.1 &11.0 &9.4 &8.1 &6.5 &9.4  \\
&AF \cite{actionformer} &\underline{26.6} &\underline{25.6} &\underline{24.4} &\underline{22.4} &\underline{18.3} &\underline{23.4}  \\
&Ours  &\textbf{28.0} &\textbf{27.2} &\textbf{25.7} &\textbf{23.7} &\textbf{20.1} &\textbf{25.0} \\
\hline
\multirow{2}{*}{Noun} &G-TAD \cite{GTAD,actionformer} &11.0 &10.0 &8.6 &7.0 &5.4 &8.4  \\
&AF \cite{actionformer} &\underline{25.5} &\underline{24.3} &\underline{22.6} &\underline{20.3} &\underline{16.6} &\underline{21.9}\\
&Ours &\textbf{26.0} &\textbf{24.4} &\textbf{23.0} &\textbf{20.4} &\textbf{16.7} &\textbf{22.1}\\ 
\hline
\end{tabular}
\end{center}
\caption{Comparative results for separate verb and noun models on the EPIC-KITCHENS-100 validation set. All methods use the same SlowFast features \cite{slowfast}. \textbf{Bold} for best model and \underline{underline} for second best.
}
\label{separate results}
\end{table*}

\begin{table*}[h!]
\scriptsize
\begin{center}
\setlength{\tabcolsep}{1.5mm}
{
\begin{tabular}{|l|l|c|cccc|c|cccccc|}
\hline
\multicolumn{1}{|c|}{\multirow {2}{*}{\bf Method}} &\multicolumn{1}{c|}{\multirow {2}{*}{\bf Venue}}  &\multicolumn{5}{c}{\bf ActivityNet-1.3} &\multicolumn{7}{|c|}{\bf THUMOS14}\\
\cline{3-14}
\multicolumn{1}{|c|}{}&\multicolumn{1}{c}{}&\multicolumn{1}{|c|}{Feature} &0.5 & 0.75 & 0.95 &Avg. &\multicolumn{1}{c|}{Feature} &0.3 & 0.4 & 0.5 & 0.6 & 0.7 &Avg.  \\ 
\hline\hline

BSN \cite{BSN} &ECCV 18 &TSN \cite{TSN} &46.5 &30.0 &8.0 &30.0 &TSN \cite{TSN} &53.5 &45.0 &36.9 &28.4 &20.0 &36.8 \\[3pt]
BMN \cite{BMN} &ICCV 19 &TSN \cite{TSN} &50.1 &34.8 &8.3 &33.9 &TSN \cite{TSN} &56.0 &47.4 &38.8 &29.7 &20.5 &38.5\\[3pt]
G-TAD \cite{GTAD}&CVPR 20 &TSP \cite{TSP} &51.3 &37.1 &9.3 &35.8 &TSN \cite{TSN} &54.5 &47.6 &40.2 &30.8 &23.4 &39.3\\[3pt]
BC-GNN \cite{BCGNN} &ECCV 20 &TSN \cite{TSN} &50.6 &34.8 &9.4 &34.3 &TSN \cite{TSN} &57.1 &49.1 &40.4 &31.2 &23.1 &40.2\\[3pt]
BSN++ \cite{BSN++} &AAAI 21 &TSN \cite{TSN} &51.3 &35.7 &8.3 &34.9 &TSN \cite{TSN} &59.9 &49.5 &41.3 &31.9 &22.8 &41.1\\[3pt]
TCANet \cite{TCANET} &CVPR 21 &SF \cite{slowfast} &54.3 &39.1 &8.4 &37.6 &TSN \cite{TSN} &60.6 &53.2 &44.6 &36.8 &26.7 &44.3 \\[3pt]
TVNet \cite{TVNet} &VISAPP 22 &TSN \cite{TSN} &51.4 &35.0 &\underline{10.1} &34.6 &TSN \cite{TSN} &64.7 &58.0 &49.3 &38.2 &26.4 &47.3\\[3pt]

DCAN \cite{DCAN} &AAAI 22 &TSN \cite{TSN} &51.8 &36.0 &9.5 &35.4 &TSN \cite{TSN} &68.2 &62.7 &54.1  &43.9 &\underline{32.6} &52.3\\[3pt]
RCL\cite{RCL} &CVPR 22 &TSP \cite{TSP} &\underline{55.2} &\underline{39.0} &8.3 &\underline{37.7} &TSN \cite{TSN} &70.1 &62.3 &52.9 &42.7 &30.7 &51.7 \\[3pt]
RefactorNet \cite{RefactorNe} &CVPR 22 &I3D \cite{I3D} &\textbf{56.6} &\textbf{40.7} &7.4 &\textbf{38.6} &I3D \cite{I3D} &\underline{70.7} &\underline{65.4} &\underline{58.6} &\underline{47.0} &32.1 &\underline{54.8} \\[3pt]

MUSES \cite{BCNet, MUSES} &AAAI 22 &I3D \cite{I3D}  &53.2 &36.2 &\textbf{10.6} &35.5 &I3D \cite{I3D} &\textbf{71.5} &\textbf{67.0} &\textbf{60.0} &\textbf{48.9} &\textbf{33.0} &\textbf{56.1}\\[3pt]
\hline\hline

SSN \cite{SSN} &ICCV 17 &TS \cite{Two-stream}  &43.3 &28.7 &5.6 &28.3 &TS \cite{Two-stream} &51.9 &41.0 &29.8 &- &- &-\\[3pt]
GTAN \cite{GTAN}&CVPR 19 &P3D \cite{P3D} &52.6 &34.1 &\underline{8.9} &34.3 &P3D \cite{P3D} &57.8 &47.2 &38.8 &- &- &- \\[3pt]
TadTR \cite{TadTR}  &TIP 22 &I3D \cite{I3D} &49.1 &32.6 &8.5 &32.3 &I3D \cite{I3D} &62.4 &57.4 &49.2 &37.8 &26.3 &46.6  \\[3pt]
AFSD \cite{AFSD} &CVPR 21 &I3D \cite{I3D} &52.4 &35.3 &6.5 &34.4 &I3D \cite{I3D} &67.3  &62.4 &55.5 &43.7 &31.1 &52.0  \\[3pt]
TALLFormer \cite{TALLFormer} &ECCV 22 &SW \cite{SWIN-B} &\underline{54.1} &36.2 &7.9 &35.6 &SW \cite{SWIN-B} &76.0 &- &63.2 &- &34.5 &59.2\\[3pt]
MRBD \cite{MRBD} &CVPR 22 &SF \cite{slowfast} &50.5 &36.0 &\textbf{10.8} &35.1  &SF \cite{slowfast} &69.4 &64.3 &56.0 &46.4 &34.9 &54.2 \\[3pt]
AF \cite{actionformer} &ECCV 22 &TSP \cite{TSP} &\underline{54.1} &36.3 &7.7 &\underline{36.0} &I3D \cite{I3D} &75.5 &72.5 &65.6 &56.6 &42.7 &62.6\\[3pt] 
AF \cite{actionformer}$^\dagger$ &ECCV 22 &TSP \cite{TSP} &\textbf{54.2} &\underline{36.9} &7.6 &\underline{36.0} &I3D \cite{I3D} &\underline{82.1}	&\underline{77.8} &\underline{71.0}	&\underline{59.4} &\underline{43.9}	&\underline{66.8}\\[3pt] 
Ours & - &TSP \cite{TSP} &\underline{54.1} &\textbf{37.3} &8.0 &\textbf{36.1}  &I3D \cite{I3D} &\textbf{82.7} &\textbf{79.0} &\textbf{71.7} &\textbf{60.9} &\textbf{46.3} &\textbf{68.1}\\[3pt] 

\hline
\end{tabular}}
\end{center}
\vspace*{-5pt}
\caption{Comparative results on ActivityNet-1.3 and THUMOS14, grouped by two-stage methods (top) and one-stage methods (bottom).
$^\dagger$ means results from latest ActionFormer codebase. 
\textbf{Bold} for best model and \underline{underline} for second best. 
}
\label{main results:AN/TH}
\end{table*}

\subsection{Results}

\noindent\textbf{Main Results:}
Tab. \ref{tab:results_epic} shows the EPIC-KITCHENS-100 results, which is the most challenging dataset we use. Our method outperforms all previous work in all mAP@IoUs. Most importantly, it surpasses the current SOTA work, i.e. ActionFormer \cite{actionformer} modified with our verb/noun classification head. This establishes a direct comparison against the the same architecture, but without boundary confidence.

Tab. \ref{separate results} shows a comparison of noun detection and verb detection separately (i.e. separate training and testing runs for each). Although this is a simpler task and not the recommended evaluation metric, it allows us to compare our method with the reported results from ActionFormer, which we also outperform.

Tab. \ref{main results:AN/TH} shows results for ActivityNet-1.3 and THUMOS14. In general, two-stage methods (top) are strong on ActivityNet-1.3 as it is a simple test, with one action per video, but our method still outperforms other one-stage methods (bottom). On the more challenging THUMOS14 dataset with multiple actions per video, one-stage methods are better. Our method achieves SOTA results using the same features as the best two-stage and one-stage methods.

\begin{figure*}[th!]
	\centering
	\includegraphics[width=0.99\linewidth]{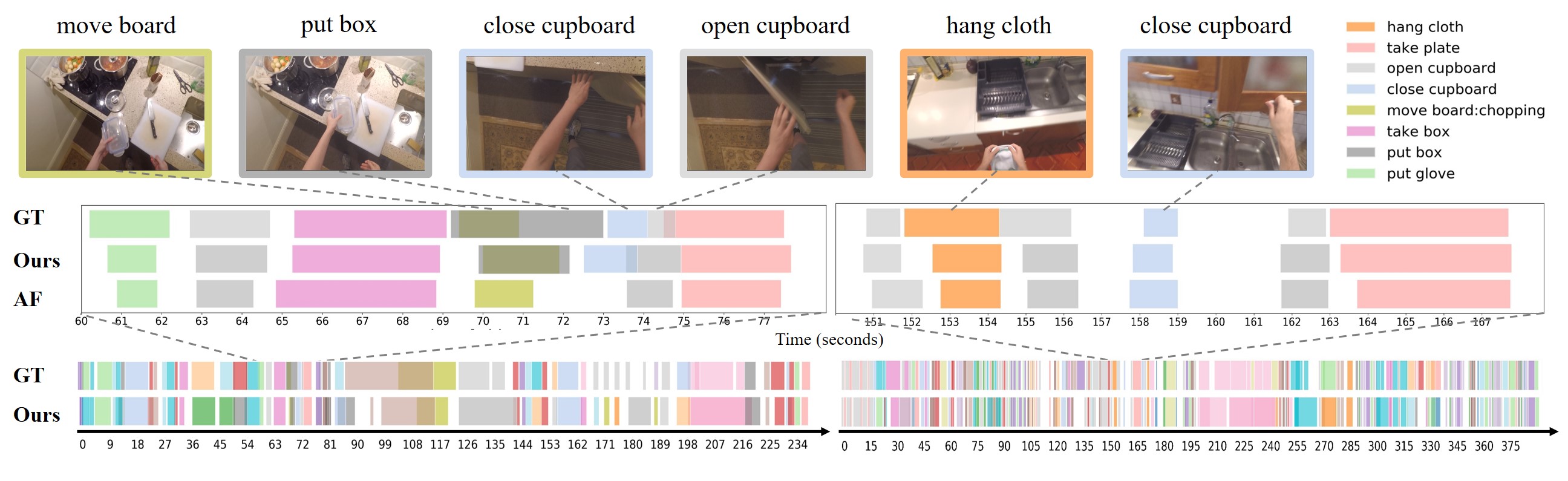}
	\vspace*{-5pt}
    \caption{Qualitative results on the EPIC-KITCHENS-100 validation set. Ground truth and predictions are shown with colour-coded class labels (see legend). 
     The bottom two lines are ground truth (GT) and our prediction (Ours) for the whole video sequence. The middle three lines show ground truth (GT), our prediction (Ours) and ActionFormer's \cite{actionformer} prediction (AF) for a zoomed-in region. The top line shows visual content of selected frames. Results demonstrate that our method is good at dense actions and boundary refinement.}
	\label{fig: qualitative}
\end{figure*}

\noindent\textbf{Qualitative Results:}
Figure \ref{fig: qualitative} shows the qualitative illustrations on EPIC-KITCHENS-100 validation set. The global results (bottom two lines) indicate the model’s capability to effectively detect dense actions with varying classes and lengths.     
The local results (middle three lines) show that our method could predict boundaries closer to the ground truth than ActionFormer, demonstrating that predicting boundary confidence is critical for one-stage action detection. 
In addition, our method captures more hard actions with overlapping than ActionFormer, such as ``close cupboard'' in the 73rd second and ``put box'' in the 72th second. ActionFormer misses these actions most likely due to imprecise predictions that lack a measure of confidence, especially for actions with similar visual content (such as frames in the top line: ``move board'' and ``put box'', ``close cupboard'' and ``open cupboard'').

\begin{table}[t]
\setlength\tabcolsep{3.5pt}
\footnotesize
\begin{minipage}[t]{1\linewidth}
\begin{center}
\begin{tabular}{|c|c|cccccc|}
\hline
\multicolumn{1}{|c}{\multirow {2}{*}{\bf Scaling}} &\multicolumn{1}{|c}{\multirow {2}{*}{\bf Task}} &\multicolumn{6}{|c|}{\bf Val (mAP@IoU)}\\
\cline{3-8}
\multicolumn{1}{|c}{}&\multicolumn{1}{|c}{}&\multicolumn{1}{|c}{0.1} & 0.2 & 0.3 & 0.4 & 0.5 & Avg. \\ 
\hline\hline
\multirow{3}{*}{\XSolidBrush}   &Verb &23.66 &22.61 &21.09 &19.02 &16.54& 20.58 \\
&Noun &22.92 &21.68 &20.30 &18.64 &16.07 &19.92\\
&Action &18.19 &17.63 &16.59 &15.38 &13.64 &16.29\\
\hline
\multirow{3}{*}{\Checkmark}   &Verb &23.75 &22.68 &22 &19.19 &16.73 &\textbf{20.71} \\
&Noun &23.58 &22.40 &21.03 &19.27 &16.39 &\textbf{20.53}\\
&Action &19.19 &18.61 &17.47 &16.30 &14.33 &\textbf{17.18}\\ 
\hline
\end{tabular}
\end{center}
\caption{
Effect of confidence tokens scaling. Without scaling (\XSolidBrush) means directly using the predicted confidence token as the boundary confidence. Scaling (\Checkmark) means scaling to boundary confidence using Equation \ref{eq:gaussian}. Results are shown for the action task in the validation set of EPIC-KITCHENS-100.
\textbf{Bold} for best.
}
\label{Ablation:scaling}
\end{minipage}
\end{table}
\begin{table}[t]
\begin{minipage}[t]{1\linewidth}
\begin{center}
\begin{tabular}{|l|cccccc|}
\hline
\multicolumn{1}{|c|}{\multirow {2}{*}{\bf $\sigma$}} &\multicolumn{6}{|c|}{\bf Val (mAP@IoU)}\\
\cline{2-7}
\multicolumn{1}{|c}{}&\multicolumn{1}{|c}{0.1} & 0.2 & 0.3 & 0.4 & 0.5 & Avg.\\ 
\hline\hline

\multirow{1}{*}{4}	&18.83 &18.33 &17.12 &15.90 &13.86 &16.81\\
\hline
\multirow{1}{*}{4.5} &18.78 &18.28 &17.19 &15.89 &14.16 &16.86\\
\hline
\multirow{1}{*}{5}	&18.96 &18.46 &17.24 &16.04 &14.31 &17.00\\
\hline
\multirow{1}{*}{5.5}	 &\textbf{19.19} &\textbf{18.61} &\textbf{17.47} &\textbf{16.30} &\textbf{14.33} &\textbf{17.18}\\
\hline
\multirow{1}{*}{6} &18.93 &18.43 & 17.22 &16.02 &14.29 &16.98\\
\hline

\end{tabular}
\end{center}
\caption{
Comparison of different values of hyper-parameters $\sigma$ in Equation \ref{eq:gaussian} of boundary head. Results are shown for the action task in the validation set of EPIC-KITCHENS-100. \textbf{Bold} for best.
}
\label{Ablation table: sigma}
\end{minipage}
\end{table}

\begin{table*}[t]
\setlength\tabcolsep{3.5pt}
\footnotesize
\begin{minipage}[t]{0.47\textwidth}
\begin{center}
\begin{tabular}{|l|cccccc|}
\hline
\multicolumn{1}{|c|}{\multirow {2}{*}{\bf Combinations}} &\multicolumn{6}{|c|}{\bf Val (mAP@IoU)}\\
\cline{2-7}
\multicolumn{1}{|c}{}&\multicolumn{1}{|c}{0.1} & 0.2 & 0.3 & 0.4 & 0.5 & Avg.\\ 
\hline\hline

\multirow{1}{*}{$\hat{p}_t^s \ast \hat{p}_t^e$}	&12.83 &11.61 &10.41 &9.03 &7.31 &10.24\\
\hline
\multirow{1}{*}{$\hat{p}_t^a$}	&18.74 &18.16 &17.23 &15.95 &13.90 &16.80\\
\hline
\multirow{1}{*}{$\hat{p}_t^a \ast \hat{p}_t^s$}	&19.02 &18.43 &17.34 &16.00 & 14.04 &16.96\\
\hline
\multirow{1}{*}{$\hat{p}_t^a \ast \hat{p}_t^e$}	&19.12 &18.57 &\textbf{17.54} &16.19 &14.23 &17.13 \\
\hline
\multirow{1}{*}{$Avg.(\hat{p}_t^a, \hat{p}_t^s, \hat{p}_t^e)$}&18.74 &18.17 &17.06 &15.60 &13.68 &16.65\\
\hline
\multirow{1}{*}{$\hat{p}_t^a \ast \hat{p}_t^s \ast \hat{p}_t^e$}	&19.18 &18.60 &17.46 &16.27 &14.28 &17.16\\
\hline
\multirow{1}{*}{$\hat{p}_t^a \ast \sqrt{\hat{p}_t^s \ast \hat{p}_t^e}$}	&\textbf{19.19} &\textbf{18.61} &17.47 &\textbf{16.30} &\textbf{14.33} &\textbf{17.18}\\
\hline
\end{tabular}
\end{center}
\caption{
Comparison of different combinations of confidence scores. Results are shown for the action task in the validation set of EPIC-KITCHENS-100.
\textbf{Bold} for best.
}
\label{Ablation table: confidence scores}
\end{minipage}
\hspace{0.6cm}
\begin{minipage}[t]{0.47\textwidth}
\begin{center}
\begin{tabular}{|c|c|cccccc|}
\hline
\multicolumn{1}{|c}{\multirow {2}{*}{\bf top-$v$}} &\multicolumn{1}{|c}{\multirow {2}{*}{\bf top-$n$}}  &\multicolumn{6}{|c|}{\bf Val (mAP@IoU)}\\
\cline{3-8}
\multicolumn{1}{|c}{}&\multicolumn{1}{|c}{}&\multicolumn{1}{|c}{0.1} & 0.2 & 0.3 & 0.4 & 0.5 & Avg. \\ 
\hline\hline
\multirow{1}{*}{1} &\multirow{1}{*}{3} &17.05  &16.47  &15.54  &14.53  &12.67  &15.25 \\
\hline
\multirow{1}{*}{2} &\multirow{1}{*}{6} &18.33  &17.82  &16.82  &15.75  &14.00  &16.54 \\
\hline
\multirow{1}{*}{5} &\multirow{1}{*}{15} &19.16 &18.57 &17.40 &16.23 &14.24 &17.12 \\
\hline
\multirow{1}{*}{10} &\multirow{1}{*}{30} &\textbf{19.19} &\textbf{18.61} &\textbf{17.47} &\textbf{16.30} &14.33 &\textbf{17.18}\\
\hline

\multirow{1}{*}{15} &\multirow{1}{*}{45} &19.16 &18.57 &17.44 &16.27 &14.30 &17.15\\
\hline

\multirow{1}{*}{20} &\multirow{1}{*}{60} &18.88  &18.35  &17.32  &16.23  &14.36  &17.03 \\
\hline

\multirow{1}{*}{97} &\multirow{1}{*}{300} &18.89 &18.36 &17.33 &16.23 &\textbf{14.37} &17.04 \\
\hline

\end{tabular}
\end{center}
\caption{Comparison of different numbers of top-$v$, top-$n$ for combining action labels. Results are shown for the action task on the validation set of EPIC-KITCHENS-100. \textbf{Bold} for best.  
}
\label{num of selected classes}
\end{minipage}
\end{table*}

\begin{figure*}[htbp]
	\centering
	\begin{minipage}{0.005\linewidth}
		\centering
		\rotatebox{90}{\scriptsize EPIC-KITCHENS-100}
	\end{minipage}
	\begin{minipage}{0.24\linewidth}
		\centering
		\includegraphics[width=1\linewidth]{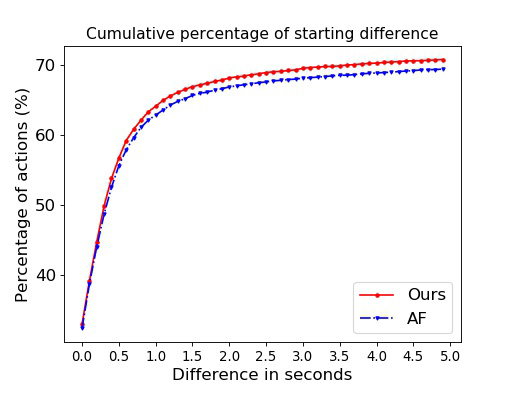}
	\end{minipage}
	\begin{minipage}{0.24\linewidth}
		\centering
		\includegraphics[width=1\linewidth]{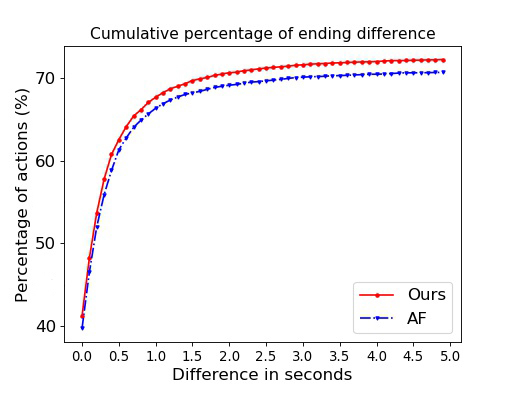}
	\end{minipage}
	\begin{minipage}{0.005\linewidth}
		\centering
		\rotatebox{90}{\scriptsize THUMOS14}
	\end{minipage}
	\begin{minipage}{0.24\linewidth}
		\centering
		\includegraphics[width=1\linewidth]{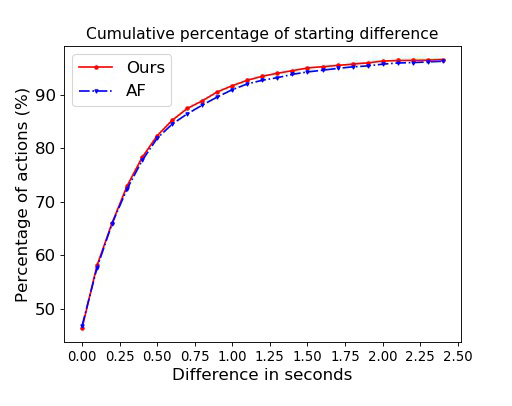}
	\end{minipage}
	\begin{minipage}{0.24\linewidth}
		\centering
		\includegraphics[width=1\linewidth]{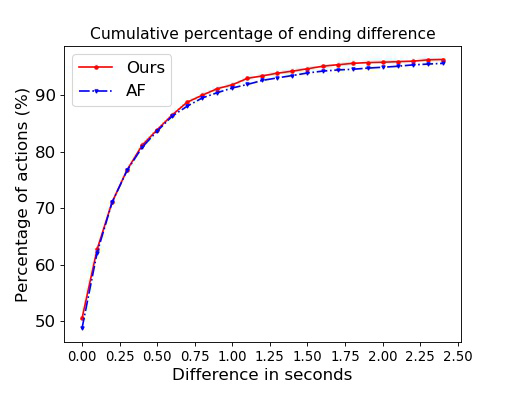}
	\end{minipage}
	\caption{Accumulative percentage of actions which is correctly detected within x seconds difference from ground truth. Acceptance difference x is shown on the x-axis. Results are shown for starting and ending on EPIC-KITCHENS-100 (left) and THUMOS14 (right) dataset. Our model outperforms ActionFormer \cite{actionformer} in all cases.}

	\label{accumulation}
\end{figure*}

\begin{figure*}[h!]
	\centering
	\includegraphics[width=0.33\linewidth]{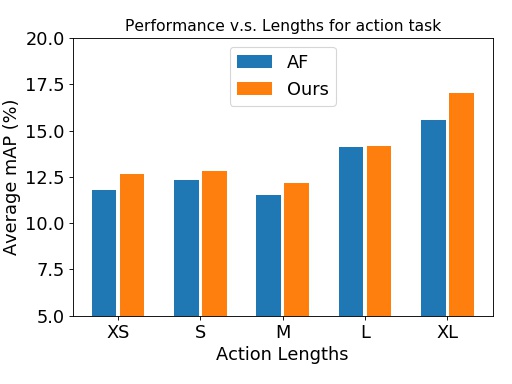}
	\includegraphics[width=0.33\linewidth]{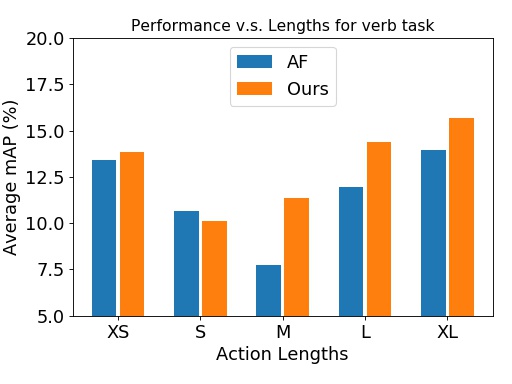}
	\includegraphics[width=0.33\linewidth]{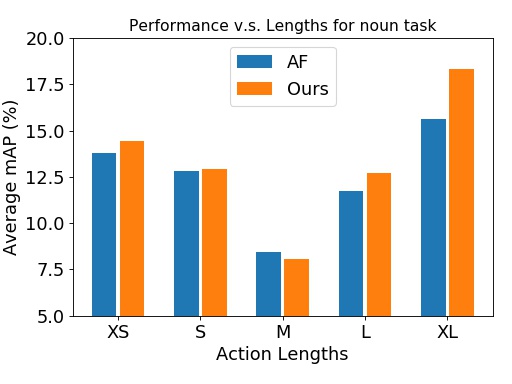}	
	\vspace*{-5pt}
    \caption{Comparing detection results to ActionFormer \cite{actionformer} at different action lengths on the validation set of EPIC-KITCHENS-100. We divide actions to five groups based on lengths (in seconds): XS (0, 2], S (2, 4], M (4, 6], L (6, 8], and XL (8, inf). Left: action task. Middle: verb task. Right: noun task.}
	\label{fig: action lengths}
\end{figure*}

\subsection{Ablations}
\label{abaltion}

\noindent\textbf{Effect of confidence scaling:}
In Section \ref{boundary head}, we assume two ways to produce the boundary confidence. The first is to directly produce boundary confidence from the top layers. The second is to produce confidence tokens and scale them to boundary confidence, which achieves better performance (see Table \ref{Ablation:scaling}). The results demonstrate that designing with scaling can better keep the consistency of predicting and optimizing, and boost the performance.

\noindent\textbf{Boundary refinement:}
To demonstrate that our boundary head can achieve more refined boundaries, we present the accumulative percentage of actions which correctly detect within a certain threshold of the boundary error. We define the boundary error as temporal difference between ground truth start/end and predicted start/end, in seconds.
As shown in Figure \ref{accumulation}, our boundary head can detect actions with more precise boundaries compared to ActionFormer \cite{actionformer} on both the EPIC-KITCHENS-100 and THUMOS14 datasets. We observe a greater improvement achieved on the EPIC-KITCHENS-100 dataset, which contains actions with a wider range of lengths, demonstrating that our method is more effective for more complex sequences. 

\noindent\textbf{Effect of sigma:}
The hyperparameter $\sigma$ in Equation \ref{eq:gaussian} determines the degree of scaling when weighting confidence tokens to output boundary confidence. Table \ref{Ablation table: sigma} shows that the results change slightly with $\sigma$ from $4$ to $6$ at a step size of $0.5$, with the best results at $\sigma=5.5$. In addition, the results show reasonable robustness to the change in $\sigma$.

\noindent\textbf{Confidence score combinations:}
Our method produces confidence scores for its three outputs: action, start boundary and end boundary. At inference, just one confidence score is required for proposal suppression. Table \ref{Ablation table: confidence scores} ablates how our three scores can be combined. We can see that the best performance is found when multiplying the action confidence by the square root of the product of the boundary confidences. Note that the boundary confidence on its own does not work, since it is hard to distinguish between whether a proposal is an action or background, and what action it is based just on the boundary.

\noindent\textbf{Effect of action length:}
A successful action detection method should perform well over a wide range of action lengths. Figure \ref{fig: action lengths} shows the average mAP on EPIC-KITCHENS-100 verb, noun and action of our method compared to ActionFormer \cite{actionformer} with our classification head, which demonstrates the effect of including boundary confidence. Although improvements are found at all action lengths, the largest improvement is on the longest actions. This is most likely due to longest actions being hard to regress given the long relative distances, but our boundary head helps to alleviate this issue because it uses the region around the boundary to compute the confidence.

\noindent\textbf{Multi-task Classification:}
On EPIC-KITCHENS-100, unlike ActionFormer, our classification head handles compound verb/noun actions by only selecting the combinations of the top-$v$ predicted verbs and top-$n$ predicted nouns. Table \ref{num of selected classes} ablates this choice. In all cases, $n=3v$ as there are three times as many noun classes compared to verb classes. Clearly just relying on very low numbers causes a significant performance penalty, as the redundant proposal suppression relies on multiple action confidences. Performance is relatively stable above $v=5$ and $n=15$. Our main results use $v=10$ and $n=30$. 

\section{Conclusion}
{In this paper, we propose a novel boundary head and incorporate it into a one-stage anchor-free pipeline to estimate boundary confidence scores and produce refined boundaries for temporal action detection.
Our method achieves state-of-the-art results on commonly used action detection benchmarks, including a significant improvement on the challenging EPIC-KITCHENS-100 dataset, which contains dense actions of various lengths.
Detailed ablations show the benefits of incorporating boundary confidences and the effect of different parameters,  establishing their importance in handling compound, egocentric actions.
In future work, we aim to explore temporal action sequence contexts and multi-modality based on language models.

\vspace{2pt}
\noindent\textbf{Acknowledgement: }Funded by the University of Bristol and the China Scholarship Council.
}

{\small
\bibliographystyle{ieee}
\bibliography{paper_ref}
}

\end{document}